\documentclass{article}

\usepackage{microtype}
\usepackage{graphicx}
\usepackage{subfigure}
\usepackage{booktabs} 
\usepackage{algorithm} 
\usepackage{algorithmic}
\usepackage[algo2e]{algorithm2e}
\usepackage{graphicx}
\usepackage{multirow}
\usepackage{makecell}
\usepackage{amsmath}

\usepackage{hyperref}

\usepackage[accepted]{icml2019}

\icmltitlerunning{Submission and Formatting Instructions for ICML 2019}

\begin{document}

\twocolumn[
\icmltitle{Toward Extremely Low Bit and Lossless Accuracy in DNNs with Progressive ADMM}






\icmlsetsymbol{equal}{*}

\begin{icmlauthorlist}
\icmlauthor{Sheng Lin}{neu}
\icmlauthor{Xiaolong Ma}{neu}
\icmlauthor{Shaokai Ye}{thu}
\icmlauthor{Geng Yuan}{neu}
\icmlauthor{Kaisheng Ma}{thu}
\icmlauthor{Yanzhi Wang}{neu}

\end{icmlauthorlist}

\icmlaffiliation{neu}{Northeastern University, Boston}
\icmlaffiliation{thu}{Tsinghua University, Beijing}

\icmlcorrespondingauthor{Sheng Lin}{lin.sheng@husky.neu.edu}
\icmlcorrespondingauthor{Xiaolong Ma}{ma.xiaol@husky.neu.edu}
\icmlcorrespondingauthor{Shaokai Ye}{shaokaiyeah@gmail.com}
\icmlcorrespondingauthor{Geng Yuan}{yuan.geng@husky.neu.edu}
\icmlcorrespondingauthor{Kaisheng Ma}{kaisheng@mail.tsinghua.edu.cn}
\icmlcorrespondingauthor{Yanzhi Wang}{yanz.wang@northeastern.edu}

\icmlkeywords{Machine Learning, ICML}

\vskip 0.3in
]



\printAffiliationsAndNotice{}  

\begin{abstract}
Weight quantization is one of the most important techniques of Deep Neural Networks (DNNs) model compression method. A recent work using systematic framework of DNN weight quantization with the advanced optimization algorithm ADMM (Alternating Direction Methods of Multipliers) achieves one of state-of-art results in weight quantization. 
In this work, we first extend such ADMM-based framework to guarantee solution feasibility and we have further developed a multi-step, progressive DNN weight quantization framework, with dual benefits of (i) achieving further weight quantization thanks to the special property of ADMM regularization, and (ii) reducing the search space within each step. Extensive experimental results demonstrate the superior performance compared with prior work.
   Some highlights: we derive the first lossless and fully binarized (for all layers) LeNet-5 for MNIST; And we derive the first fully binarized (for all layers) VGG-16 for CIFAR-10 and ResNet for ImageNet with reasonable accuracy loss. Our models and sample codes are released in anonymous link  \url{http://bit.ly/2YYqzJv}.
\end{abstract}

\section{Introduction}
With the development of machine learning technologies, Deep Neural Networks (DNNs) have shown their extraordinary performance for their high accuracy and excellent scalability\cite{krizhevsky2012imagenet}. However, DNNs are suffering from both intensive computation and huge storage. A number of prior work have focused on developing \emph{model compression} techniques for DNNs. These techniques, which are applied during the training phase of the DNN, aim to simultaneously reduce the model size and accelerate the computation for inference -- all these to be achieved with non-negligible classification accuracy loss. Indeed the accuracy of a DNN inference engine after model compression is typically higher than that of a shallow neural network with no compression \cite{han2015learning, wen2016learning}. 
One of the most important categories of DNN model compression techniques is \emph{weight quantization}.

We have investigated weight quantization of DNNs in many recent work \cite{leng2017extremely,park2017weighted,zhou2017incremental,lin2016fixed,wu2016quantized,rastegari2016xnor,hubara2016binarized,courbariaux2015binaryconnect}. In these work, both storage and computational requirements of DNNs have been greatly reduced with tolerable accuracy loss. We know that multiplication operations are costly and it can be eliminated when applying binary, ternary, or power-of-2 weight quantizations \cite{rastegari2016xnor,hubara2016binarized,courbariaux2015binaryconnect}.

To overcome the limitation of the highly heuristic nature in prior work, a recent work \cite{leng2017extremely} developed a systematic framework of DNN weight quantization using the advanced optimization technique ADMM \cite{boyd2011distributed,hong2016convergence}. Through the adoption of ADMM, the original weight quantization problem is decomposed into two sub-problems, one effectively solved using stochastic gradient descent as original DNN training, while the other solved optimally and analytically via Euclidean projection. This method achieves one of state-of-art in weight quantization results. However, the direct application of ADMM technique lacks the guarantee on solution feasibility due to the non-convex nature of objective function (loss function), while there is also margin of improvement for solution quality.

In this work, we first make the following extensions on the ADMM-based weight compression \cite{zhang2018systematic}: (i) develop an integrated framework of dynamic ADMM regularization and quantized weight projection, thereby guaranteeing solution feasibility and providing high solution quality; (ii) incorporate the multi-$\rho$ updating technique for faster and better ADMM convergence.

Extensive experimental results demonstrate that the proposed progressive framework consistently outperforms prior work. Some highlights: we derive the first lossless, fully binarized (for all layers) LeNet-5 for MNIST; and we derive fully binarized (for all layers) VGG-16 model for CIFAR-10 and ResNet model for ImageNet with reasonable accuracy loss.

\section{DNN Model Compression}
In this section, we give a detailed description to achieve a good quantization result for DNNs with progressive ADMM. 
\subsection{Framework Design}

\begin{figure} [!h]
     \centering
     \includegraphics[width=0.8\columnwidth]{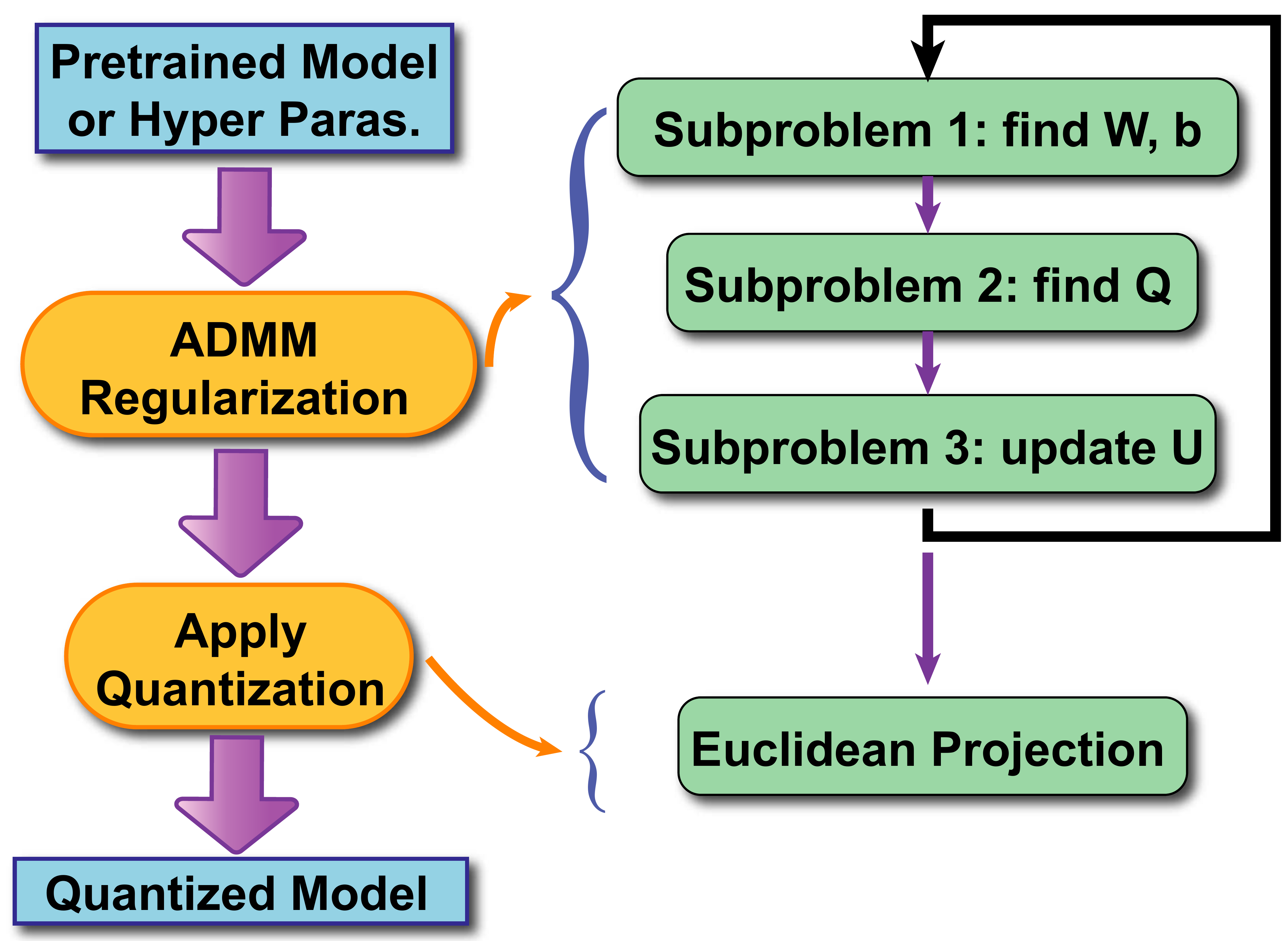}  
     \caption{Illustration of one step progressive DNN model quantization.}
     \label{fig:general_step}
 \end{figure}

The ADMM-based weight quantization is performed multiple times, each as a step in the progressive framework. Figure \ref{fig:general_step} illustrates one step of proposed progressive DNN weight quantization framework. The quantization result from the previous step is evaluated with current quantization result, and serve as intermediate results and starting point for the subsequent step if it is better than current result. The reason to develop a progressive model compression framework is that the multi-step procedure reduces the search space for weight quantization within each step.  

Through extensive investigations, we conclude that the progressive comparison will be in general sufficient for weight quantization, in which each step requires approximately the same number of training epochs. And during the process, we adjust the penalty factor of ADMM to speed up the convergence and achieve better quantized result. The specific ADMM optimization process is introduced in following subsection.  

\subsection{ADMM-based Weight Quantization}

ADMM\cite{boyd2011distributed} is an advanced optimization technique which decompose an original problem into sub-problems that can be solved separately and iteratively. By adopting ADMM regularized optimization, the framework can provide high solution quality and with no obvious accuracy degradation.

First, the {\textit{progressive DNN weight quantization}} starts from a pre-trained full size DNN model without compression. Consider an $N$-layer DNNs, sets of weights of the $i$-th (CONV or FC) layer are denoted by ${\bf{W}}_{i}$, respectively. And the \textit{loss function} associated with the DNN is denoted by $f \big( \{{\bf{W}}_{i}\}_{i=1}^N \big)$. In this paper, $\{{\bf{W}}_{i}\}_{i=1}^N$ characterize the set of weights from layer $1$ to layer $N$.
 The overall weight quantization problem is defined by
\begin{equation}
\small
\label{original}
\begin{aligned}
& \underset{ \{{\bf{W}}_{i}\}}{\text{minimize}}
& & f \big( \{{\bf{W}}_{i}\}_{i=1}^N \big),
\\ & \text{subject to}
& & {\bf{W}}_{i}\in {\bf{\mathcal{Q}}}_{i}, \; i = 1, \ldots, N.
\end{aligned}
\end{equation}
For weight quantization, elements in ${\bf{\mathcal{Q}}}_{i}$ are the solutions of ${\bf{W}}_{i}$. Assume set of ${q_{i,1}, q_{i,2}, \cdots, q_{i,M_{i}}}$ is the available quantized values, where $M_i$ denotes the number of available quantization level in layer $i$. Suppose $q_{i,j}$ indicates the $j$-th quantization level in layer $i$, which gives
\begin{equation}
    q_{i,j}\in{\{-\alpha_{i},\alpha_{i}\}} \: or \: {\{ -\alpha_{i},0,\alpha_{i}\}}.
\end{equation}
And ${\alpha_{i}}$ is the scaling factor, which is initialized by the average of weight values in layer $i$.

Then the original problem (\ref{original}) can be equivalently rewritten as 
\begin{equation}
\small
\label{admm_form}
\begin{aligned}
& \underset{ \{{\bf{W}}_{i}\}}{\text{minimize}}
& & f \big( \{{\bf{W}}_{i} \}_{i=1}^N \big)+\sum_{i=1}^{N} h_{i}({\bf{Q}}_{i}),
\\ & \text{subject to}
& & {\bf{W}}_{i} = \textbf{Q}_i, \; i = 1, \ldots, N.
\end{aligned}
\end{equation}

We incorporate auxiliary variables ${\bf{Q}}_{i}$, dual variables ${\bf{U}}_{i}$, then apply ADMM to decompose problem (\ref{admm_form}) into simpler subproblems. Then solve these subproblems iteratively until convergence. The augmented Lagrangian formation of problem (\ref{admm_form}) is
\vspace{-0.50em}
\begin{equation}
\small
\begin{aligned}
\label{equ7}
 \underset{ \{{\bf{W}}_{i}\} \}}{\text{minimize}}
\ \ \ & f \big( \{{\bf{W}}_{i} \}_{i=1}^N \big) +  \sum_{i=1}^{N} \frac{\rho_{i}}{2}  \| {\bf{W}}_{i}-{\bf{Q}}_{i}+{\bf{U}}_{i} \|_{F}^{2} \\
\end{aligned}
\end{equation}

The first term in problem (\ref{equ7}) is the differentiable loss function of the DNN, and the second term is a quadratic regularization term of the ${\bf{W}}_{i}$, which is differentiable and convex. As a result, subproblem (\ref{equ7}) can be solved by stochastic gradient descent algorithm~\cite{kingma2014adam} as the original DNN training.

The standard ADMM algorithm~\cite{boyd2011distributed} steps proceed by repeating, for $k = 0, 1,\ldots$, the following subproblems iterations:
\vspace{-0.20em}
\begin{equation}
\small
    \bf{W}_{i}^{k+1} := \underset{ {\bf{W}}_{i}}{\text{arg min}}\quad \textit{L}_p(\{\bf{W}_i\}, \{\bf{Q}_i^k\}, \{\bf{U}_i^k\}),
\label{itera1}
\end{equation}
\vspace{-0.50em}
\begin{equation}
\small
    \bf{Q}_{i}^{k+1} := \underset{ {\bf{Q}}_{i}}{\text{arg min}}\ \textit{L}_p(\{\bf{W}_i^{k+1}\}, \{\bf{Q}_i\}, \{\bf{U}_i^k\}),
\label{itera2}
\end{equation}
\vspace{-0.50em}
\begin{equation}
\small
    \bf{U}_{i}^{k+1} := \bf{U}_{i}^{k}+\bf{W}_{i}^{k+1}-\bf{Q}_{i}^{k+1}.
\label{itera3}
\end{equation}

which (\ref{itera1}) is the proximal step, (\ref{itera2}) is projection step and (\ref{itera3}) is dual variables update.

\section{Experimental Results}
\emph{\textbf{Binary Weight Quantization Results on LeNet-5}}: To the extent of authors' knowledge, we achieve the first lossless, fully binarized LeNet-5 model in which weights all layers are binarized. The accuracy is still 99.21\%, lossless compared with baseline. 
For example, recent work \cite{cheng2018differentiable} results in 2.3\% accuracy degradation on MNIST for full binarization, with baseline accuracy 98.66\%.

\begin{table}[ht]
\centering
\caption{Comparisons of fully binary weight quantization results on LeNet-5 for MINIST dataset.}\label{table:Lenet-quan}
\begin{tabular}{p{2.5cm}p{1.5cm}p{2.5cm}}
\hline
Method & Accuracy &  Num. of bits \\ 
\hline
Baseline \cite{cheng2018differentiable} & 98.66\% &  32 \\
\hline
 Binary \cite{cheng2018differentiable}  & 96.34\% &  1 \\  \hline
\bf{Our binary}  & 99.21\% &  1 \\ \hline 
\end{tabular}
\end{table}

\emph{\textbf{Weight Quantization on CIFAR-10}}: We also achieve fully binarized VGG-16 for CIFAR-10 with negligible loss in accuracy, in which weights all layers (including the first and the last) are binarized. The accuracy is 93.58\%. We would like to point out that fully ternarized quantization results in 94.02\% accuracy. Table \ref{table:VGG-quan} shows our results and comparisons.

\begin{table}[ht]
\centering
\caption{Comparisons of fully binary (ternary) weight quantization results on VGG-16 for CIFAR-10 dataset.}\label{table:VGG-quan}
\begin{tabular}{p{2.5cm}p{1.5cm}p{2.5cm}}
\hline
Method & Accuracy &  Num. of bits \\ 
\hline
Baseline \cite{cheng2018differentiable} & 84.80\% &  32 \\ \hline
 8-bit \cite{cheng2018differentiable} & 84.07\% &  8 \\ \hline
 Binary \cite{cheng2018differentiable}  & 81.56\% &  1 \\ \hline
 \bf{Our baseline}  & 94.70 \% &  32 \\ \hline
 \bf{Our ternary}  & 94.02\% &  2  \\ \hline
\bf{Our binary}  & 93.58\% &  1 \\ \hline 
\end{tabular}
\end{table}

\emph{\textbf{Binary Weight Quantization Results on ResNet for ImageNet Dataset}}: 
The binarization of ResNet models on ImageNet data set is widely acknowledged as a very challenging task. As a result, there are very limited prior work (e.g., the one-shot ADMM \cite{leng2017extremely}) with binarization results on ResNet models. As \cite{leng2017extremely} targets ResNet-18 (which is even more challenging than ResNet-50 or larger ones), we make a fair comparison on the same model. Table \ref{table:ResNet-quan} demonstrates the comparison results (Top-5 accuracy loss).
  In prior work, it is by default that the first and last layers are not quantified (or quantized to 8 bits) as these layers have a significant effect on overall accuracy. When leaving the first and last layers unquantized, our framework is not progressive, but an extended one-shot ADMM-based framework. We can observe the higher accuracy compared with the prior method under this circumstance (first and last layers unquantized while the rest of layers binarized).
The Top-1 accuracy has similar result: 3.8\% degradation in our extended one-shot and 4.3\% in \cite{leng2017extremely}.

\begin{table}[ht]
\centering
\caption{Comparisons of weight quantization results on ResNet-18 for ImageNet dataset.}\label{table:ResNet-quan}
\begin{tabular}{p{2.7cm}p{2cm}p{2.2cm}}
\hline
Method & Relative Top-5 acc. loss  & Num. of bits \\ 
\hline
Uncompressed & 0.0\%  & 32 \\ \hline
 {One-shot ADMM quantization \cite{leng2017extremely}}  & 2.9\%  & 1 (32 for the first and last) \\ 
\hline
\bf{Our method}  & 2.5\%  & 1 (32 for the first and last)  \\ \hline
\bf{Our method}  & 5.8\%  & 1 \\ \hline
\end{tabular}
\end{table}

Using the progressive framework, we can derive a fully binarized ResNet-18, in which weights in all layers are binarized. The accuracy degradation is 5.8\%, which is noticeable and shows that the full binarization of ResNet is a challenging task even under the progressive framework. We did not find prior work for comparison on this result.

\section{Conclusion and On-going Work}
In this work, we extended the prior ADMM-based framework and developed a multi-step, progressive DNN weight quantization framework, in which we achieve further weight quantization results and provide better convergence rate. 

Considering the good performance of our method, we plan to test more different networks for ImageNet dataset. And we are working on testing our method for different applications and datasets.


\end{document}